\documentclass[conference]{IEEEtran}
\makeatletter\if@twocolumn\PassOptionsToPackage{switch}{lineno}\else\fi\makeatother
\usepackage{cite}
\usepackage{graphicx}
\usepackage{amsmath,amssymb,amsfonts}
\usepackage{algorithmic}
\usepackage{graphicx}
\usepackage{textcomp}
\usepackage{xcolor}
\usepackage[numbers,sort&compress]{natbib}
\usepackage{graphicx}
\usepackage{threeparttable}
\usepackage{multirow}
\usepackage{lmodern}
\usepackage{tcolorbox}
\usepackage{caption}
\usepackage{booktabs}
\usepackage{lipsum}
\usepackage{tabularx} 
\usepackage{adjustbox}
\usepackage{etoolbox}
\usepackage{bigdelim}

\usepackage{graphicx}
\usepackage[T1]{fontenc}
\usepackage[numbers,sort&compress]{natbib}
\usepackage{mathtools}
\usepackage{booktabs}
\usepackage{lineno,hyperref}
\usepackage{amsmath}
\usepackage{amsfonts}
\usepackage{amssymb}
\usepackage{amsmath}
\usepackage{multirow,morefloats,floatflt,cancel,tfrupee}
\usepackage{makecell}
\usepackage{tikz}
\usepackage{setspace}
\usepackage{color}
\usepackage{multirow}
\usepackage{bbding}
\usepackage{physics}
\usepackage{colortbl}
\usepackage{xcolor}
\usepackage{pifont}
\usepackage[nointegrals]{wasysym}
\renewcommand\IEEEkeywordsname{Keywords}

\newcommand{\teblebold}[1]{\small\textbf{#1}}

\makeatletter
\def\ps@IEEEtitlepagestyle{%
  \def\@oddfoot{\mycopyrightnotice}%
  \def\@evenfoot{}%
}
\def\mycopyrightnotice{%
  {\footnotesize 979-8-3503-3015-1/23/\$31.00  \copyright 2023 IEEE \hfill}
  \gdef\mycopyrightnotice{}
}

\let\old@ps@IEEEtitlepagestyle\ps@IEEEtitlepagestyle
\def\confheader#1{%
    \def\ps@IEEEtitlepagestyle{%
        \old@ps@IEEEtitlepagestyle%
        \def\@oddhead{\strut\hfill#1\hfill\strut}%
        \def\@evenhead{\strut\hfill#1\hfill\strut}%
    }%
    \ps@headings%
}
\makeatother

\confheader{%
    \parbox{17cm}{\centering 13th International Conference on Computer and Knowledge Engineering (ICCKE 2023), November 1-2, 2023, Ferdowsi University of Mashhad}
}

\begin{document}

\title{Efficient Vision Transformer for Accurate Traffic Sign Detection}

\makeatletter
\newcommand{\linebreakand}{%
  \end{@IEEEauthorhalign}
  \hfill\mbox{}\par
  \mbox{}\hfill\begin{@IEEEauthorhalign}
}


\author{Javad Mirzapour Kaleybar$^{1}$
\quad Hooman Khaloo$^{2}$
\quad Avaz Naghipour$^{1}$
\\
${^1}$ Department of Computer Engineering, University College of Nabi Akram, Tabriz, Iran  \\
${^2}$ School of Technology Sharif University, Tehran, Iran\\
}


\maketitle

\begin{abstract}
This research paper addresses the challenges associated with traffic sign detection in self-driving vehicles and driver assistance systems. The development of reliable and highly accurate algorithms is crucial for the widespread adoption of traffic sign recognition and detection (TSRD) in diverse real-life scenarios. However, this task is complicated by suboptimal traffic images affected by factors such as camera movement, adverse weather conditions, and inadequate lighting.
This study specifically focuses on traffic sign detection methods and introduces the application of the Transformer model, particularly the Vision Transformer variants, to tackle this task. The Transformer's attention mechanism, originally designed for natural language processing, offers improved parallel efficiency. Vision Transformers have demonstrated success in various domains, including autonomous driving, object detection, healthcare, and defense-related applications. To enhance the efficiency of the Transformer model, the research proposes a novel strategy that integrates a locality inductive bias and a transformer module. This includes the introduction of the Efficient Convolution Block and the Local Transformer Block, which effectively capture short-term and long-term dependency information, thereby improving both detection speed and accuracy. Experimental evaluations demonstrate the significant advancements achieved by this approach, particularly when applied to the GTSDB dataset.
\end{abstract}

\begin{IEEEkeywords}
Vision Transformer; Object Detection; Traffic Sign Detection; Auto-Driving Cars
\end{IEEEkeywords}

\begin{figure*}
	\centering
	\includegraphics[width=.6\textwidth]{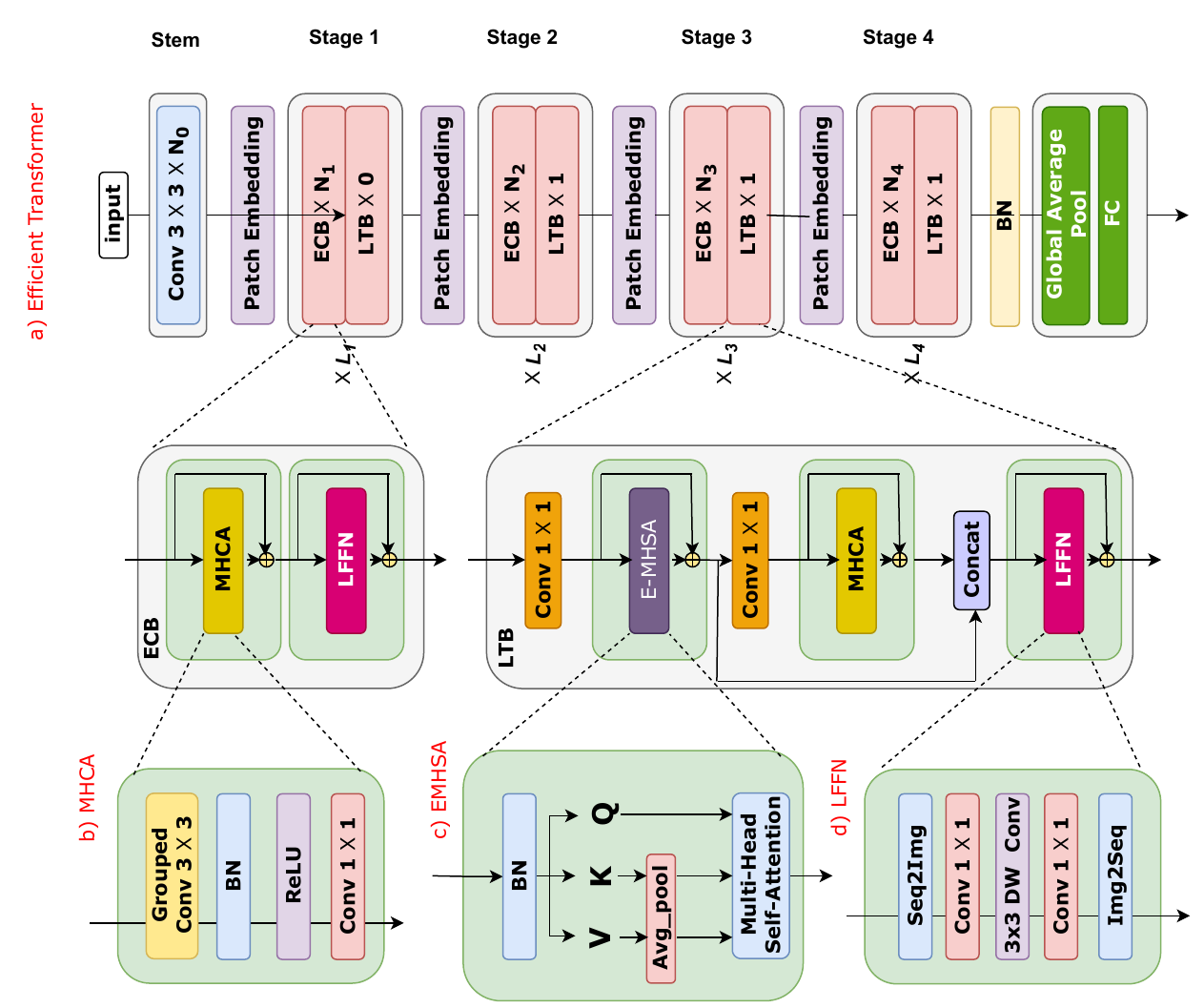}
    \caption{\textbf{Network structure of the presented Efficient Vision Transformer.}}
	\label{FIG:Architecture}
\end{figure*}

\section{Introduction}

Driver assistance systems and autonomous vehicles rely heavily on traffic sign recognition and detection (TSRD). An essential requirement for the extensive adoption of TSRD is the development of an algorithm that is both dependable and highly accurate in diverse real-life situations. Furthermore, traffic images captured along the road are typically less than optimal, as well as the vast collection of traffic signs that must be identified. These images frequently suffer from distortions caused by camera movement, unfavorable weather conditions, and inadequate lighting, all of which substantially amplify the challenges associated with this task.

It is an area of research that remains complex, and a variety of investigations have been conducted in this field over the years. A comprehensive collection of these studies is provided in \cite{ellahyani2021traffic}. TSRD involves two distinct tasks: Traffic Sign Recognition (TSR) and Traffic Sign Detection (TSD). Conventionally, TSD methods heavily relied on manual feature extraction in order to extract features from the source dataset. A manual algorithm was used to extract features based on color information, geometric shapes, and edge detection. Color-based approaches typically involve segmenting traffic sign regions in specific color spaces including Hue-Saturation-Intensity (HSI, Hue-Chroma-Luminance (HCL), and others. Although these solutions necessitated significant efforts to manually curate large volumes of images, which proved to be expensive and time-consuming. Shape- and color-based methods were also widely employed but exhibited common weaknesses such as sensitivity to illumination changes, occlusions, scale variations, rotations, and translations. Although Machine Learning (ML) can address some of these issues, it requires extensive annotated datasets. over the past years, Deep Learning (DL) has revealed as a powerful methodology to traffic sign recognition, achieving state-of-the-art (SOTA) performance. DL research has led to the development of various models, including ResNet~\cite{Resnet}, VGG16, and MobileNet.  A second important factor contributing to deep learning success is the use of open datasets in previous studies, including the German Traffic Sign Detection Benchmark (GTSDB)~\cite{GTSDB}.

This study specifically focuses on TSD methods. The authors have to the best of their knowledge not been able to find any previous study that has applied the Transform model specifically to the detection of traffic signs. A key aspect of Transformer's architecture is the attention mechanism, which was created for addressing sequence-to-sequence challenges in the NLP. Unlike traditional recurrent neural networks (RNNs) like LSTM or GRU, as well as convolutional neural networks (CNNs), the Transformer solely relies on attention operations, omitting recursion or convolution. This design improves parallel efficiency while maintaining performance. Vision Transformers have been introduced in the computer vision domain as innovative solutions to visual tasks. There are now a number of real-world applications for these networks, including autonomous driving~\cite{zhang2022cctsdb}, object detection~\cite{pyramid}, health care~\cite{medvit, Dilated}, and defense-related applications~\cite{Embeded}. To be precise, this study employs for Vision Transformer variants: the Pyramid Vision Transformer model (PVT)~\cite{PVT}, Swin Transformer~\cite{liu2021swin}, Locality iN Locality (LNL)~\cite{LNL}, and Pyramid Transformer~\cite{pyramid}. As baseline model, ResNet~\cite{Resnet} is also utilized.

In this research paper, a novel strategy is suggested to strengthen the efficiency of the Transformer model by integrating locality inductive bias and a transformer module. The approach involves the combination of two fundamental blocks. Firstly, the Efficient Convolution Block (ECB) is introduced, which is specifically designed to capture short-term dependency information in visual data. This is achieved through the utilization of a unique and uncomplicated configuration Multi-Head Convolutional Attention (MHCA) mechanism. Furthermore, the Local Transformer Block (LTB) is introduced, which excels in capturing high frequency data information while also functioning as a portable signal mixer for both high and low-frequency signals. This mixing capability enhances the overall modeling capacity of the system. Notably, each block comprises of two parallel forks: convolution layers and one comprising local self-attention, and the other supplemented by a feed-forward network (LFFN). Experimental evaluations demonstrate the significant improvements achieved by this approach in metrics of accuracy and detection speed, particularly when applied to the GTSDB dataset.

\begin{table*}[t]
    \caption{The performance of the SOTA models in detecting objects of GTSDB}
    \label{Tab:comparision}
    \begin{adjustbox}{width=.7\textwidth,center}

        \begin{tabular}{c |c| c| c c c}
            \toprule
            Decoder & backbone & params (M) & $AP$ & $AP^{75}$ & $AP^{50}$ \\
            \midrule
            & ResNet-50 \cite{Resnet}& 44 & 63.4 & 67.3 & 83.1 \\
            & LNL-T \cite{LNL}& 46 & 68.3 & 72.7 & 91.1 \\
Faster RCNN & Swin-T \cite{liu2021swin}& 48 & 68.4 & 72.6 & 90.6 \\
            & PVT-S \cite{PVT}& 44 & 65.4 & 67.3 & 87.7 \\
            & Pyramid \cite{pyramid}& 36 & 70.2 & 74.6 & 91.2 \\
            & Ours & 32 & \teblebold{72.3} & \teblebold{75.8} & \teblebold{93.5} \\
            \midrule
            & ResNet-50 \cite{Resnet}& 82 & 70.7 & 74.6 & 88.4\\
            & LNL-T \cite{LNL}& 86 & 75.4 & 79.9 & 94.2 \\
Cascade RCNN & Swin-T \cite{liu2021swinSwin}& 86 & 74.5 & 78.8 & 93.3 \\
            & PVT-T \cite{PVT}& 83 & 75.2 & 79.4 & 93.9\\
            & Pyramid \cite{pyramid}& 74 & 77.8 & 81.8 & 96.5 \\
            & Ours & 71 & \teblebold{78.3} & \teblebold{83.7} & \teblebold{97.0} \\
            \bottomrule
        \end{tabular}
    \end{adjustbox}
\end{table*}

\section{METHOD}

Our goal is to create a novel approach that combines the transformer block and convolution layers, resulting in an efficient hybrid architecture specifically designed for detecting traffic sign images. As depicted in Figure~\ref{FIG:Architecture}, our architecture consists of transformer blocks, a patch od embed layers, and a series of stacked convolutions in all phases, hierarchically follows by the established pyramid structure. During this process, the dimensional resolution decreases gradually with a full ratio of $32\times$, achieved through reductions of [$4\times$, $2\times$, $2\times$, and $2\times$], in every stage of the convolution process, the channel size is doubled after each convolution block. In this section, our primary objective is to examine the primary blocks charge for embedding multi-scale features. Subsequently, we improve the capturing of high level and low level data information in the input data by developing enhanced versions of the LTB and ECB.

\subsection{Revisit vision transformer}

Our approach builds upon the Vision Transformer (ViT) framework, and thus, we present a concise summary of ViT \cite{Vit}. As opposed to the commonly used convolutional neural network (CNN) methods for classification task, ViT employs a fully attention-based approach. Within each transformer architecture of ViT, two fundamental components are employed: a feed-forward network (FFN), multi-head self-attention (MSA), featuring layer normalization and residual skip connections. ViT separates pictures into fixed-sized tokens, such as $17\times17$, then conducts linear projection to convert them into tokens in order to modify transformers for vision-related tasks. These patch tokens are then combined with class tokens to create the input sequence. We include a trainable absolute positional embedding that is introduced before the transformer encoders get the input in order to collect positional information for each token. Finally, the class token is utilized as the ultimate feature description at the conclusion of the network. The ViT framework can be summarized as bellow:

\begin{align}
   &&x_0 &= [x_{patch}||x_{cls}] + x_{pos},\\
   &&y_k &= x_{k-1} + MSA(LN(x_{k-1})),  \\
   &&x_k &= y_k + FFN(LN(y_k)),
\end{align}

In the provided equation, $ x_{cls} \in R^{1\times C} $ corresponds to the class tokens, $ x_{patch} \in R^{N\times C} $ represents the patch tokens, and $ x_{pos} \in R^{(1+N)\times C} $ denotes the position embeddings. Within this context, $C$ signifies the amount of patch tokens, $K$ denotes the layer indicator, and $N$ represents the embedding size. Nevertheless, the vanilla ViT model, despite its ability to learn global interactions between all patch tokens, encounters memory efficiency issues with self-attention. This is due to the quadratic growth in computational cost as the number of tokens increases. Consequently, the standard ViT model proves inadequate for vision tasks that necessitate high-resolution details, such as object detection. In order to address these limitations, we introduce the pyramid module for the vision transformer.

\subsection{Efficient Convolutional Block}

To enhance the robustness and accuracy of traffic sign detection, we propose an Efficient Convolution Block (ECB) that demonstrates exceptional performing as a hybrid transformer block, while also maintaining the deployment advantages of the residual convolutional block. The ECB adopts a hybrid structure, that has been established as essential for leveraging multi frequency data information, as depicted in Figure~\ref{FIG:Architecture}. Simultaneously, an efficient attention-based token combining component plays a crucial role. We devise a LFFN as a means of introducing locality into the architecture, employing depth-wise convolution, and a MHCA as an effective token combiner. Taking inspiration from the Next Vision Transformer~\cite{medvit}, which analyzed the impact of each Transformer component on robust detection, we construct the ECB by mixing the MHCA and LFFN blocks within the robust design. The suggested ECB could be arranged as bellow:

\begin{equation}
    \begin{aligned}
        \tilde{z}^{l} &=\operatorname{MHCA}\left(z^{l-1}\right)+z^{l-1} \\
        z^{l} &=\operatorname{LFFN}\left(\tilde{z}^{l}\right)+\tilde{z}^{l}
        \end{aligned}
\end{equation}

In the upcoming section, we will provide a thorough explanation of LFNN, while referring to the input from the $l-1$ block as $z^{l-1}$, and denoting the outputs of MHCA and the $l$ ECB as $\tilde{z}^{l}$ and $z^l$ respectively.

\subsubsection{LFFN structure}

The application of the FFN to the sequence of tokens $\mathbf{Z}^{r}$ involves rearranging them into a 2D lattice, as illustrated in Figure~\ref{FIG:Architecture} (d). This rearrangement enables a precise depiction of the reshaped features, which can be explained as follows:

\begin{equation}
\mathbf{Z}^{r}=\operatorname{Seq} 2 \operatorname{Img}(\mathbf{Z}), \mathbf{Z}^{r} \in \mathbb{R}^{h \times w \times d}
\end{equation}

Given the condition $w=W/p$ and $h=H/p$, the function $Seq2IMG$ is employed to convert a sequence into a attribute space that can be effectively demonstrated. One token is assigned to each pixel on the feature map during this transformation, signifying one token per pixel. With the use of this technology, closeness between tokens may be reinstated, introducing localization into the network. Consequently, it becomes possible to substitute the fully-connected layers with $1 \times 1$ convolutional layers, as demonstrated below:

\begin{equation}
\begin{aligned}
\mathbf{Y}^{r} &=f\left(\mathbf{Z}^{r} \circledast \mathbf{W}_{1}^{r}\right) \circledast \mathbf{W}_{2}^{r} \\
\mathbf{Y} &=\operatorname{Img} 2 \mathrm{Seq}\left(\mathbf{Y}^{r}\right)
\end{aligned}
\end{equation}

The kernels of the convolutional layers, $\mathbf{W}{1}^{r} \in \mathbb{R}^{d \times \gamma d \times 1 \times 1}$ and $\mathbf{W}{2}^{r} \in \mathbb{R}^{{\gamma} d \times d \times 1 \times 1}$, are modified versions of $W_1$ and $W_2$ respectively. These kernels play a crucial role in the convolutional layers. Afterwards, the image feature map is regenerated back into a token series using the $Img2Seq$ function. The resulting token sequence is then transformed into the fused token, which is subsequently utilized by the following self-attention layer.

\subsection{Local Transformer Block}

Considering previous shortcoming, we have suggested the Local Transformer Block (LTB) with the objective of efficiently capturing multi-frequency data in a lightweight manner. The LTB acts as an effective mixer for multi-frequency signals, enhancing the whole modeling capability of the network. Figure~\ref{FIG:Architecture}~(c) illustrates the LTB, wherein it begins by capturing low-frequency signals through the utilization of an Efficient Self Attention (ESA). The formulation of ESA can be represented as follows:

\begin{equation}
\begin{aligned}
\operatorname{ESA}(x) &=\operatorname{Concat}\left(\operatorname{SA}_{1}\left(x_{1}\right), \mathrm{SA}_{2}\left(x_{2}\right), \ldots, \mathrm{SA}_{h}\left(x_{h}\right)\right) W^{O} \\
\operatorname{SA}(X) &=\operatorname{Attention}\left(X \cdot W^{Q}, \mathrm{P}_{s}\left(X \cdot W^{K}\right), \mathrm{P}_{s}\left(X \cdot W^{V}\right)\right)
\end{aligned}
\end{equation}

We utilize the notation $X = [x_1, x_2, ..., x_h]$ to indicate the division of the input feature $X$ into a multi-head form along the channel dimension. The output projection layer is denoted as $W^O$, and the variable $h$ represents the number of heads. We developed SA based on the linear SRA approach mentioned in \cite{PVT} to reduce the spatial resolution of self-attention.
The attention calculation involves the linear layers $W^Q, W^K, W^V$, following the standard attention form: $Attention(Q,K,V) = \text{softmax}(QK^T / \sqrt{d})V$, where $d$ stands for the transformer's hidden spacial size. The $Ps$ operation employs an average-pool operation with a stride parameter $s$ to downsample the spatial dimensions before applying the attention operation. This downsampling is performed to reduce computational costs. Additionally, we have observed that the number of channels in the ESA module significantly influences the time consumption of the module.

It is crucial to emphasize that LTB integrates a multi-frequency configuration that collaborates with ESA and MHCA modules. Additionally, we introduce a novel attention mechanism in LTB that exploits efficient convolutional operations to improve its efficiency. Inspired by the successful multi-head design in MHSA, we develop our convolutional attention (CA) using a multi-head paradigm. This enables the CA to concurrently attend to information from different representation subspaces at various positions, effectively facilitating local representation learning. The formulation of the proposed MHCA is as follows:

\begin{equation}
\begin{aligned}
\operatorname{MHCA}(x) &=\operatorname{Concat}\left(\mathrm{CA}_1\left(x_1\right), \mathrm{CA}_2\left(x_2\right), \ldots, \mathrm{CA}_h\left(x_h\right)\right) W^O \\
\mathrm{CA}(X) &= \left(\mathrm{W} \cdot \mathrm{T}_{\{i,j\}}\right), \text { where } \mathrm{T}_{\{i,j\}} \in X
\end{aligned}
\end{equation}

In the proposed formulation, the MHCA module is responsible for capturing information from h parallel representation subspaces, whereas CA represents the single-head convolutional attention. The calculation involves the trainable parameter $W$ and the adjacent tokens $\mathrm{T}{{i,j}}$ within the input vector $X$. The opration CA is determined by performing an inner product function between the rearranged vectors of the nearby tokens $\mathrm{T}{{i,j}}$ and the learnable parameter $W$. While the MHSA in Transformers is effective in capturing global context, our proposed CA, derived from MHCA, specifically focuses on learning the relationship between different tokens within the local receptive field. It is important to note that our implementation of MHCA combines both point-wise convolution and group convolution (multi-head convolution), as depicted in Figure~\ref{FIG:Architecture}~(b).

\section{EXPERIMENTS and RESULTS}

In this section, we commence by offering a summary of the datasets employed in training our Efficient Transformer. Subsequently, we elucidate the experimental settings that were utilized. Lastly, we present a thorough analysis of the experimental results obtained from our Transformer model, particularly focusing on its performance in traffic sign detection. Additionally, we compare these results with the SOTA studies in the field.

\subsection{Evaluation Measures and Dataset}\label{AA}

To assess the performance of our model, we employ the mean average precision (mAP) as the standard evaluation metric commonly utilized in object detection datasets. The average precision (AP) quantifies the tradeoff between recall and precision by calculating the area under the recall-precision curve. We compute the metric AP for each class of objects and then determine the average value to obtain the mAP score for all classes within the dataset. Additionally, we utilize the Intersection-over-Union (IoU) metric, specifically using IoU thresholds of 0.5 and 0.75, to further assess the mAP.

Our proposed model is subjected to evaluation using the GTSDB dataset, which is selected due to its widespread usage and reputation as a benchmark for comparing traffic sign detection methods. The GTSDB dataset comprises natural traffic scenes captured under diverse weather conditions (rain, sunny, and fog) and on numerous kinds of roads (highway, rural, and urban) during both daylight and twilight. It encompasses a total of 900 full images containing 1,206 traffic signs, with a division into a train subset consisting of 631 images (856 traffic signs) and a test subset containing 323 images (370 traffic signs). The images in this dataset exhibit different characteristics, including scenarios with one, zero, or multiple traffic signs, as well as challenges such as varying orientations, occlusions, and low-light environment. The 43 classes in the GTSDB dataset are divided into the superclasses including danger, prohibitory, mandatory, and other. To conduct our evaluation, we focus on a subset of the 43 traffic sign classes within the GTSDB, as traffic signs within the similar superclass may convey distinct data, such as different speed limits (e.g., 'speed limit 80' vs. 'speed limit 30').

\subsection{Implementation details}

We examine the effectiveness of our Efficient Transformer by integrating it into two widely recognized meta-architectures such as Cascade RCNN \cite{cai2018cascade} and Faster RCNN. Our proposed structure serves as the backbone for these frameworks. To provide fair comparisons, we follow the same training approach as described in \cite{liu2021swin} and strive to maintain compatibility with related works.

Table~\ref{Tab:comparision} presents the detection results on the GTSDB dataset \cite{GTSDB} using a standard schedule of 30 epochs. The Faster RCNN model is trained with a mini-batch of 12 over 30 epochs, while employing the AdamW optimizer. We initialize the learning rate to $0.0001$ and include a warmup period of 500 iterations. At epochs $8$ and $13$, the initial learning rate is declined by a ratio factor of $0.05$. A weight decay of $0.00015$ and a uniformity loss regularization of $0.3$ are applied. We utilize the MMDetection toolboxes for implementation. Following the guidelines in \cite{liu2021swin}, the Cascade RCNN model is trained for 3x, by an initial learning rate of $0.0003$ and a weight decay of $0.0006$. All other adjustment factors align with the default settings used in Pyramid \cite{pyramid}. To ensure diversity during training, the input images are resized randomly such that the longer spatial dimension is lower than 1300 pixels, while the shorter spatial dimension falls around 840 and 400, as part of the multi-train scheme.

\subsection{Comparison with the SOTA models}

The results displayed in Table~\ref{Tab:comparision} provide compelling evidence of the exceptional performance achieved by the Efficient Vision Transformer, despite its remarkably low parameter count. The successful outcomes can be attributed to the incorporation of the ECB module and the integration of injected inductive bias. When employing the decoder of Faster RCNN in the 30-epoch configuration, our proposed model exhibits significant enhancements compared to Pyramid \cite{pyramid}, with notable improvements of 2.3\% in $AP^{50}$, 1.2\% in $AP^{75}$, and 1.1\% in overall $AP$. Furthermore, our model's efficient structural design delivers substantial performance advantages when compared to other backbones such as PVT and  ResNet. Notably, in the case of utilizing the decoder of Cascade RCNN, our method achieves the highest mAP as indicated in Table~\ref{Tab:comparision}. Impressively, with only 30 epochs of training, we achieve outstanding results, with $78.3\%$ $AP$, $83.7\%$ $AP^{75}$, and $97.0\%$ $AP^{50}$, surpassing the performance of both Cascade RCNN and Faster RCNN by 2.1\% and 2.4\% in mAP, respectively. The superior performance of the Efficient Transformer can be assigned to the introduction of inductive bias in the Transformer Block, enabling our model to effectively leverage the available data and further enhance object detection capabilities.

\section{CONCLUSION}

In conclusion, this research emphasizes the importance of developing dependable and highly accurate algorithms for TSRD in driver assistance systems and self driving cars. The complex nature of this task, compounded by suboptimal traffic images affected by various factors, presents significant challenges. This study introduces the novel application of the Transformer model, specifically Vision Transformer variants, to traffic sign detection. The proposed strategy integrates locality inductive bias and a transformer module, resulting in significant improvements in detection speed and accuracy. Experimental evaluations validate the success of our architecture, particularly on the GTSDB dataset. Future investigations can further delve into the potential of Transformer-based methods in advancing TSRD technologies.



\hyphenpenalty=10000 
\bibliographystyle{IEEEtran} 				
\bibliography{cas-refs}

\begin{thebibliography}{10}
\providecommand{\url}[1]{#1}
\csname url@samestyle\endcsname
\providecommand{\newblock}{\relax}
\providecommand{\bibinfo}[2]{#2}
\providecommand{\BIBentrySTDinterwordspacing}{\spaceskip=0pt\relax}
\providecommand{\BIBentryALTinterwordstretchfactor}{4}
\providecommand{\BIBentryALTinterwordspacing}{\spaceskip=\fontdimen2\font plus
\BIBentryALTinterwordstretchfactor\fontdimen3\font minus
  \fontdimen4\font\relax}
\providecommand{\BIBforeignlanguage}[2]{{%
\expandafter\ifx\csname l@#1\endcsname\relax
\typeout{** WARNING: IEEEtran.bst: No hyphenation pattern has been}%
\typeout{** loaded for the language `#1'. Using the pattern for}%
\typeout{** the default language instead.}%
\else
\language=\csname l@#1\endcsname
\fi
#2}}
\providecommand{\BIBdecl}{\relax}
\BIBdecl

\bibitem{ellahyani2021traffic}
A.~Ellahyani, I.~El~Jaafari, and S.~Charfi, ``Traffic sign detection for
  intelligent transportation systems: a survey,'' in \emph{E3S web of
  conferences}, vol. 229.\hskip 1em plus 0.5em minus 0.4em\relax EDP Sciences,
  2021, p. 01006.

\bibitem{Resnet}
M.~Rahimzadeh and A.~Attar, ``A modified deep convolutional neural network for
  detecting covid-19 and pneumonia from chest x-ray images based on the
  concatenation of xception and resnet50v2,'' \emph{Informatics in medicine
  unlocked}, vol.~19, p. 100360, 2020.

\bibitem{GTSDB}
J.~Stallkamp, M.~Schlipsing, J.~Salmen, and C.~Igel, ``Man vs. computer:
  Benchmarking machine learning algorithms for traffic sign recognition,''
  \emph{Neural networks}, vol.~32, pp. 323--332, 2012.

\bibitem{zhang2022cctsdb}
J.~Zhang, X.~Zou, L.-D. Kuang, J.~Wang, R.~S. Sherratt, and X.~Yu, ``Cctsdb
  2021: a more comprehensive traffic sign detection benchmark,''
  \emph{Human-centric Computing and Information Sciences}, vol.~12, 2022.

\bibitem{pyramid}
O.~N. Manzari, A.~Boudesh, and S.~B. Shokouhi, ``Pyramid transformer for
  traffic sign detection,'' in \emph{2022 12th International Conference on
  Computer and Knowledge Engineering (ICCKE)}.\hskip 1em plus 0.5em minus
  0.4em\relax IEEE, 2022, pp. 112--116.

\bibitem{medvit}
O.~N. Manzari, H.~Ahmadabadi, H.~Kashiani, S.~B. Shokouhi, and A.~Ayatollahi,
  ``Medvit: A robust vision transformer for generalized medical image
  classification,'' \emph{Computers in Biology and Medicine}, vol. 157, p.
  106791, 2023.

\bibitem{Dilated}
D.~Saadati, O.~N. Manzari, and S.~Mirzakuchaki, ``Dilated-unet: A fast and
  accurate medical image segmentation approach using a dilated transformer and
  u-net architecture,'' \emph{arXiv preprint arXiv:2304.11450}, 2023.

\bibitem{Embeded}
O.~N. Manzari and S.~B. Shokouhi, ``A robust network for embedded traffic sign
  recognition,'' in \emph{2021 11th International Conference on Computer
  Engineering and Knowledge (ICCKE)}.\hskip 1em plus 0.5em minus 0.4em\relax
  IEEE, 2021, pp. 447--451.

\bibitem{PVT}
W.~Wang, E.~Xie, X.~Li, D.-P. Fan, K.~Song, D.~Liang, T.~Lu, P.~Luo, and
  L.~Shao, ``Pyramid vision transformer: A versatile backbone for dense
  prediction without convolutions,'' in \emph{Proceedings of the IEEE/CVF
  international conference on computer vision}, 2021, pp. 568--578.

\bibitem{liu2021swin}
Z.~Liu, Y.~Lin, Y.~Cao, H.~Hu, Y.~Wei, Z.~Zhang, S.~Lin, and B.~Guo, ``Swin
  transformer: Hierarchical vision transformer using shifted windows,'' in
  \emph{Proceedings of the IEEE/CVF international conference on computer
  vision}, 2021, pp. 10\,012--10\,022.

\bibitem{LNL}
O.~N. Manzari, H.~Kashiani, H.~A. Dehkordi, and S.~B. Shokouhi, ``Robust
  transformer with locality inductive bias and feature normalization,''
  \emph{Engineering Science and Technology, an International Journal}, vol.~38,
  p. 101320, 2023.

\bibitem{Vit}
A.~Dosovitskiy, L.~Beyer, A.~Kolesnikov, D.~Weissenborn, X.~Zhai,
  T.~Unterthiner, M.~Dehghani, M.~Minderer, G.~Heigold, S.~Gelly \emph{et~al.},
  ``An image is worth 16x16 words: Transformers for image recognition at
  scale,'' \emph{arXiv preprint arXiv:2010.11929}, 2020.

\bibitem{cai2018cascade}
Z.~Cai and N.~Vasconcelos, ``Cascade r-cnn: Delving into high quality object
  detection,'' in \emph{Proceedings of the IEEE conference on computer vision
  and pattern recognition}, 2018, pp. 6154--6162.

\end{thebibliography}
\vskip8pt

\end{document}